\documentclass{article}
\usepackage{ijcai26}

\usepackage{times}
\usepackage{soul}
\usepackage{url}
\usepackage[hidelinks]{hyperref}
\hypersetup{
  pdftitle={Retrieval-Warmed Energy-Based Reasoning: A Five-Arm Ablation Methodology for Diffusion-as-Inference on Structured Reasoning Tasks},
  pdfauthor={Libo Sun, Po-Wei Harn, Zewei Zhang, Peixiong He, Xiao Qin}
}
\usepackage[utf8]{inputenc}
\usepackage[small]{caption}
\usepackage{graphicx}
\usepackage{amsmath}
\usepackage{amssymb}
\usepackage{amsthm}
\usepackage{booktabs}
\usepackage{algorithm}
\usepackage{algorithmic}
\usepackage[pagewise,switch]{lineno}

\graphicspath{{figures/}}

\title{Retrieval-Warmed Energy-Based Reasoning: A Five-Arm Ablation
Methodology for Diffusion-as-Inference on Structured Reasoning Tasks}

\author{
Libo Sun$^1$\and
Po-Wei Harn$^2$\and
Zewei Zhang$^1$\and
Peixiong He$^1$\and
Xiao Qin$^{1,\dagger}$\\
\affiliations
$^1$Department of Computer Science and Software Engineering, Auburn University, Auburn, AL 36830, USA\\
$^2$Department of Information Management, National Central University, Taoyuan 320317, Taiwan\\
$^\dagger$Corresponding author.\\
\emails
libo@auburn.edu,
harnpowei@ncu.edu.tw,
zez0001@auburn.edu,
pzh0029@auburn.edu,
xqin@auburn.edu
}

\begin{document}

\maketitle

\begin{abstract}
Warm-started diffusion samplers accelerate iterative inference, but
it is rarely clear which part of the pipeline carries the gain. We
study \textbf{retrieval-warmed energy-based reasoning (RW-EBR)} ---
an IRED energy-based diffusion model \cite{du2024ired} augmented
with a Modern Hopfield trajectory memory --- and contribute a
\textbf{five-arm ablation methodology} (oracle, best-constant,
per-query-random, shuffled, aligned) that separates three confounded
effects: class-prior bias shift, stochastic warm-starting, and
graph-aligned value reuse. The diagnostic decomposition is adapted from
LLM-RAG evaluation \cite{ru2024ragchecker}. On \textbf{connectivity-2}
(Erd\H{o}s--R\'enyi all-pairs reachability), the
aligned-vs-shuffled-oracle swing reaches \textbf{$+35$\,pp} balanced
accuracy on a fixed 1{,}000-graph validation-set diagnostic, with
value distribution and retrieval mechanics fixed, only per-graph
alignment destroyed, while per-query random initialisation falls
below cold --- per-graph alignment, not bias shift or stochasticity,
dominates. Yet the \emph{deployable} cold-prediction pipeline misses
the acceptance gate at stored-value quality. The same diagnostic logic, stopped at the
key-quality screen, applied to \textbf{Sudoku} with a task-specific
key encoder produces a clean negative at a \emph{different}
component --- key quality, under the current setup. The
decomposition names the first blocking component on each task. The
setting --- graph reachability refined by an iterative diffusion
sampler, with explainability of failure modes as the lens --- places
the work within structured and spatio-temporal reasoning.
\end{abstract}

\section{Introduction}

Iterative inference procedures --- diffusion samplers, energy-based
reasoning models --- are increasingly \emph{warm-started}: rather than
initialising from noise, the sampler is seeded with a candidate
solution, often one retrieved from a memory of past solutions, to cut
the number of refinement steps. When such a pipeline improves, or
fails, it is rarely clear \emph{which} part is responsible. A
warm-start can help because the retrieved content is genuinely
task-relevant, because it shifts the initialisation toward a better
region regardless of content, or simply because any per-query
perturbation breaks a degenerate equilibrium. These explanations imply
very different things about when retrieval warm-starting will
generalise, yet a single end-to-end accuracy number cannot tell them
apart. We frame this as explainable failure attribution for
structured-reasoning systems: localising which component of a reasoning
pipeline drives an outcome is a question of diagnostic evaluation, not
of aggregate benchmark performance. The setting that grounds the
study --- relational structure (all-pairs reachability over an
Erd\H{o}s--R\'enyi graph) refined by an iterative diffusion
sampler --- is a graph- and iteration-shaped instance of
structured and spatio-temporal reasoning.

We study this attribution problem in \textbf{retrieval-warmed
energy-based reasoning (RW-EBR)}: an IRED energy-based diffusion model
\cite{du2024ired} augmented with a Modern Hopfield trajectory memory
\cite{ramsauer2021hopfield} that supplies a per-query warm-start. Our
contribution is a \textbf{five-arm ablation methodology} --- oracle,
best-constant, per-query-random, shuffled, and aligned --- that
separates three confounded effects of a retrieval warm-start:
class-prior bias shift, stochastic warm-starting, and graph-aligned
value reuse. We organise the analysis with a three-component
decomposition --- key quality, warm-start mechanism, stored-value
quality --- adapting the retriever- and generator-side diagnostic logic
of LLM-RAG evaluation \cite{ru2024ragchecker}. We claim neither the
decomposition nor the partial-noise warm-start mechanism as novel ---
SDEdit \cite{meng2022sdedit} is the mechanism's predecessor, with WSD
\cite{scholz2025wsd} the closest learned-warm-start competitor. The
contribution is the ablation methodology and its application to
retrieval-warmed iterative inference, plus the two empirical findings
it surfaces.

The \textbf{first finding} is an alignment effect on connectivity-2
(all-pairs reachability on Erd\H{o}s--R\'enyi graphs). Under oracle
memory, the swing between the aligned arm and the shuffled arm --- gold
values whose (key, value) pairings are permuted across queries, holding
the value distribution and retrieval mechanics fixed --- is
\textbf{$+35$\,pp} in balanced accuracy. A constant-init sweep bounds
the bias-shift contribution to $\le +8$\,pp, and per-query random
initialisation lands at $-1.5$ to $-3.1$\,pp: per-graph alignment, not
bias shift and not per-query stochasticity, is the dominant lever.
These warm-start arms are run as a fixed validation-set diagnostic
(1{,}000 graphs, seed 20260420); multi-seed warm-start replication
is left for a larger study. The
deployable cold-prediction pipeline nonetheless misses the $-2$\,pp
acceptance gate ($\Delta$bal $= -4.09$\,pp); the same decomposition
localises that failure to stored-value quality.

The \textbf{second finding} is that the failure mode is heterogeneous.
We apply the same diagnostic logic --- contrastive key training and
the quality-ratio gate, stopped at the key-quality screen --- to
Sudoku with a task-specific encoder, and obtain a clean negative at a
\emph{different} component: under the current mask-aware
solved-board target and 500-candidate pool, the key encoder itself
cannot clear its quality gate, whereas on connectivity the encoder
passes and stored-value quality is the bottleneck. The two case studies
show the three-component decomposition surfacing a different
bottleneck on each task, rather than collapsing them into a single
end-to-end number.

We wrap IRED as the base reasoning model and do not modify it: IRED's
sampler initialises from a Gaussian (its Algorithm~2 hard-codes
$\tilde y \sim \mathcal{N}(0, I)$). Our principled negatives
concern the retrieval addition we study, not the IRED backbone.

In summary, we contribute (i) a five-arm ablation methodology for
retrieval-warmed iterative inference, separating bias shift, stochastic
warm-starting, and aligned value reuse; (ii) an
aligned-vs-shuffled-oracle alignment effect on connectivity-2 that
isolates per-graph value alignment as the dominant lever; and (iii) a
heterogeneous-failure case study --- under the same diagnostic workflow,
connectivity-2 fails at stored-value quality and Sudoku at key
quality --- identifying the first blocking component on each
task.

\begin{figure*}[t]
\centering
\includegraphics[width=\textwidth]{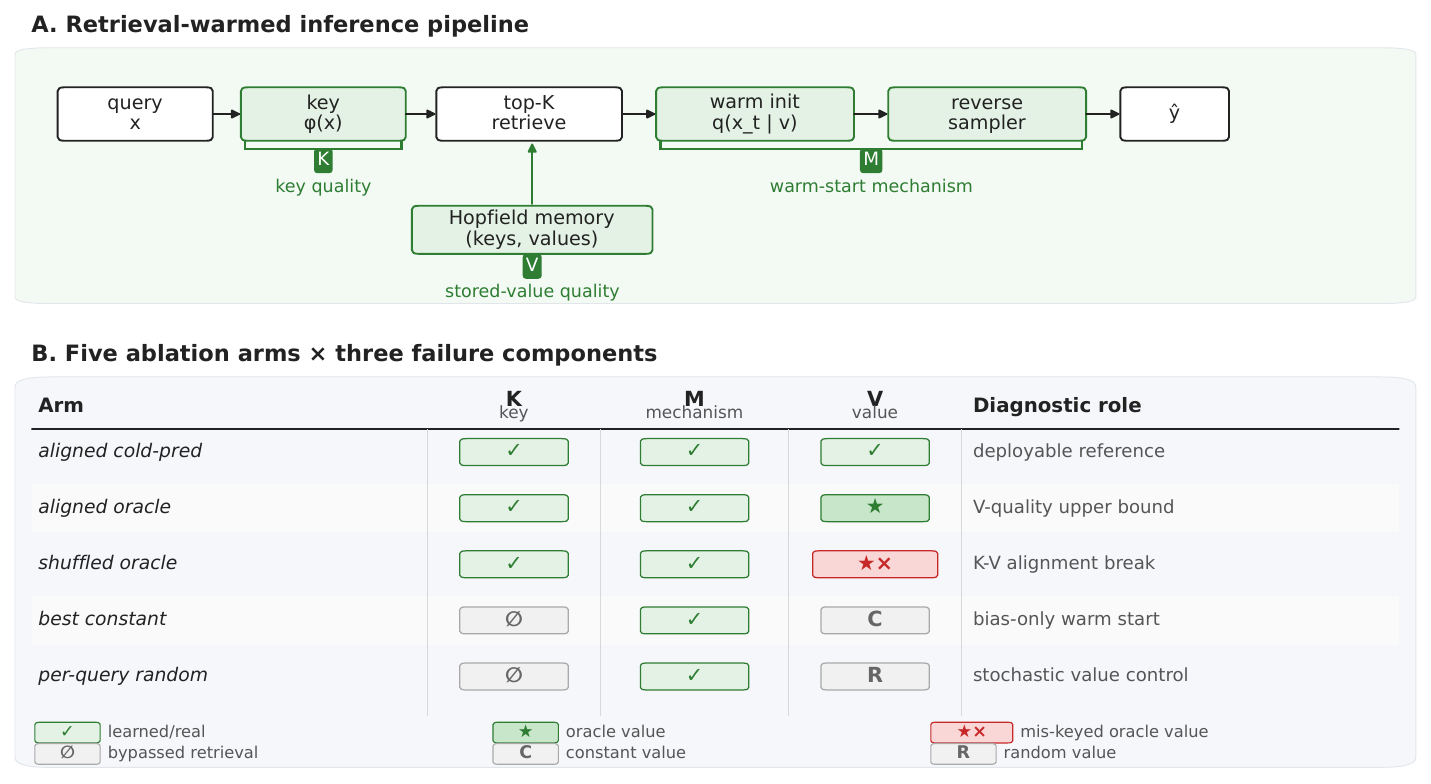}
\caption{\textbf{Diagnostic apparatus for retrieval-warmed inference.}
(A) RW-EBR pipeline annotated with the three testable components:
key quality (K), warm-start mechanism (M), stored-value quality (V).
(B) Five-arm suite as a component matrix: \checkmark{} = learned/real,
$\star$ = oracle stored values, $\star{\times}$ = mis-keyed oracle,
$\emptyset$ = bypassed task-informative retrieval, C/R = constant or random stored
values.}
\label{fig:f2}
\end{figure*}

\section{Methods}

We evaluate on \textbf{connectivity-2} \cite{du2024ired}: predict the
all-pairs reachability matrix of an undirected Erd\H{o}s--R\'enyi
$G(N{=}12, p{=}0.2)$ graph from its adjacency. Adjacency and targets
are rescaled to $\pm 1$; this is cell-level binary classification with
positive (``path exists'') prior $\approx 0.63$. Training and
evaluation both use IRED's \texttt{GraphConnectivityDataset}. The base
reasoning model (\textbf{G0}) is IRED's 32-channel GraphEBM, trained
for 30k steps under the IRED denoising objective (MSE on rescaled
$\pm 1$ targets) at 10 diffusion timesteps, with energy-landscape
supervision and inner-loop optimisation enabled. A sample drawn at $T$
inference timesteps from the unaugmented model is the
\textbf{cold $T{=}T$} baseline.

Retrieval keys come from a 3-layer GIN \cite{xu2019gin} with a
\textbf{label-ordered readout} --- per-node features concatenated in
node-label order rather than sum-pooled --- and a learned per-position
id embedding on the layer-0 features. Label-ordered readout is
engineering for our setting: sum-pool readouts are
permutation-invariant, which would destroy the label-indexed structure
the warm-start consumer needs, since retrieved reachability matrices
must align with the query's node labelling; the id embedding
distinguishes degree-symmetric collapse cases. The encoder is trained
with a supervised contrastive loss \cite{khosla2020supcon} on the 4
target-nearest neighbours of each anchor at temperature $\tau{=}0.1$
for 3{,}000 steps; the pair-similarity target is per-edge Hamming
agreement of reachability. GIN and SupCon are off-the-shelf and the
design choices are engineering, not a contribution;
Section~\ref{sec:c1} reports the resulting key quality.

A capacity-10{,}000 Modern Hopfield trajectory memory
\cite{ramsauer2021hopfield} stores (key, value) pairs whose values are
base-model trajectories, populated \textbf{write-once} during a warm-up
phase by running cold inference at $T_{\mathrm{anchor}}{=}10$ over a
random stream of 10{,}000 training examples; there are no eval-time
writes. At eval time a query's key retrieves a value via a
$\beta$-temperature softmax over the top-8 cosine similarities, where
the inverse temperature $\beta$ controls retrieval peakedness, and the
retrieved value seeds the IRED sampler in one of two ways.
\textbf{Option~A} replaces the $t{=}0$ initialisation with the
retrieved value and runs $K_{\mathrm{refine}}$ optimisation-step
iterations. \textbf{Option~B} --- the primary reported path ---
forward-noises the retrieved value to an injection timestep
$t_{\mathrm{inject}}$ via the standard diffusion forward marginal
$q(x_t \mid x_0)$, then runs the reverse IRED \texttt{p\_sample\_loop}
from $t_{\mathrm{inject}}$ down to 0. Reported runs use
$t_{\mathrm{inject}}{=}2$ --- a mild re-noising on the model's
10-timestep diffusion schedule --- with $\beta{=}20$; Option~A at
$K_{\mathrm{refine}}{=}10$ is reported alongside it for robustness.
Both reduce forward-pass count relative to a full cold sample.

All connectivity G0 and G1 runs share a fixed cached validation set of
1{,}000 graphs (seed 20260420). The headline metric is
\textbf{balanced accuracy} $= \tfrac{1}{2}(\mathrm{rec}_{+} +
\mathrm{rec}_{-})$, reported with raw accuracy and per-class recall;
the $\approx 63/37$ class imbalance makes raw accuracy a misleading
gate, whereas balanced accuracy is immune to prior-collapse on either
class. The \textbf{G1 acceptance gate} requires
$\Delta\mathrm{bal\_acc}(\mathrm{warm} - \mathrm{cold}) \ge -2$\,pp at
a forward-pass speedup $\ge 2\times$. We use this gate as an
\emph{operational diagnostic screen} for the present study --- a
sanity threshold for when to stop reporting an arm as a candidate
deployable warm-start, not a claim about external task-level
success. The $-2$\,pp tolerance is roughly $15\times$ the per-seed cold
noise floor (balanced-accuracy std $0.13$\,pp across 5 seeds;
Section~\ref{sec:c3}): any violation lies well outside per-seed
sampling jitter, while still permitting a small accuracy drop when
offset by the speedup. PASS/FAIL labels below refer to this internal
screen.

We additionally exercise component~K on Sudoku using the SATNet-style
\cite{wang2019satnet} dataset from IRED \cite{du2024ired}. The key encoder is a 3-layer
ResNet ($\sim$593k parameters) trained with the same SupCon objective
for 3{,}000 steps; the per-anchor similarity target is per-cell argmax
agreement restricted to query unknowns, scored against a 500-candidate
in-batch pool. Pass criteria are $\texttt{quality\_ratio} \ge 0.85$
and $\texttt{ret\_top\_w}(\beta{=}20) \ge 0.30$. The warm-start
mechanism and stored-value components were not exercised on Sudoku;
Section~\ref{sec:sudoku} reports the result.

\paragraph{Reproducibility.}
Upon publication we plan to release a supplementary archive containing
training and evaluation scripts for the connectivity-2 G0
backbone, the contrastive key encoder, and all five G1 warm-start
arms (cold, oracle, shuffled, best-constant, and per-query random),
together with their fixed configurations; the Sudoku key-encoder
training script and SupCon configuration; the validation-set seed
and the 5-seed cold noise-floor script; and the figure scripts
producing all six figures of the paper. The archive reproduces all
reported tables and figures against the cached validation set, and
includes a unit-test suite covering the evaluation utilities.

\section{Decomposing Retrieval-Warmed Inference}
\label{sec:decomp}

We decompose retrieval-warmed inference into three components that can
fail independently: (K) \textbf{key quality} --- does the encoder map
inputs to keys whose retrieved values warm-start usefully? Measured by
\texttt{quality\_ratio}: the target similarity between a query and its
top-1 retrieved candidate, divided by the best target similarity
available among all candidates; (M) \textbf{warm-start
mechanism} --- given a retrieved value of fixed quality, does the
inference loop refine it toward the true target? Isolated by an
\textbf{oracle-memory ablation} writing ground-truth values directly to
memory; (V) \textbf{stored-value quality} --- does the cold model
produce predictions useful as future warm-starts? This adapts LLM-RAG
diagnostic decompositions \cite{ru2024ragchecker,ragx2026} to
retrieval-warmed \emph{iterative} inference. Figure~\ref{fig:f2}
summarises the apparatus and the five arms.

\begin{table}[!t]
\centering
\small
\setlength{\tabcolsep}{4pt}
\begin{tabular}{lrrrrr}
\toprule
arm / config & bal & $r_{+}$ & $r_{-}$ & spd & $\Delta$bal \\
\midrule
\multicolumn{6}{l}{\emph{Sanity arm — deployable cold-pred memory:}} \\
cold ($T{=}10$)                 & .755 & 1.000 & .511 & $1.0\times$ & --- \\
\textbf{warm, cold-pred Opt~B}  & \textbf{.715} & .999 & \textbf{.430} & $3.3\times$ & \textbf{$-4.09$} \\
\midrule
\multicolumn{6}{l}{\emph{Oracle arm --- ground-truth memory:}} \\
cold ($T{=}10$)                 & .753 & 1.000 & .505 & $1.0\times$ & --- \\
\textbf{warm, oracle Opt~B}     & \textbf{.977} & .993 & \textbf{.960} & $3.3\times$ & \textbf{$+22.39$} \\
\textbf{warm, oracle Opt~A}     & .957 & .989 & .926 & $5.5\times$ & $+20.45$ \\
\bottomrule
\end{tabular}
\caption{Warm-start arms on connectivity-2 ($\beta{=}20$; Opt~B at
$t_{\mathrm{inj}}{=}2$, Opt~A at $K_{\mathrm{ref}}{=}10$). Sanity
arm: deployable cold-pred memory misses the gate at stored-value
quality (V FAIL). Oracle arm: both warm-start variants clear it
(M PASS). Within-arm $\Delta$bal vs each arm's own cold reference.
Fixed $n{=}1{,}000$ validation set; multi-seed caveat in
§\ref{sec:why}.}
\label{tab:warm}
\end{table}

\section{Connectivity-2: A Stored-Value-Quality Failure}
\label{sec:conn}

We walk the three components in turn on connectivity-2: the key encoder
(Section~\ref{sec:c1}), the warm-start mechanism
(Section~\ref{sec:c2}), and stored-value quality
(Section~\ref{sec:c3}). The five-arm ablation suite then decomposes the
oracle result (Section~\ref{sec:decompose-lift}), and two further
interventions characterise the failure (Section~\ref{sec:why}).

\subsection{Key Quality}
\label{sec:c1}

Validation \texttt{quality\_ratio} saturates at $\approx 0.95$
($\texttt{gt\_pred\_top1} \approx 0.83$, $\texttt{gt\_best} \approx
0.88$); $\texttt{ret\_top\_w}(\beta{=}20)$ --- the peakedness of
softmax retrieval weights --- reaches 0.49 against the $1/8 = 0.125$
uniform floor (a prior MLP+MSE baseline plateaued at 0.40). The encoder
passes the targeted relabeled-isomorph regression test --- two
labelings of a 4-node perfect matching that share all-degree-1 nodes
yet whose reachability matrices agree on only 50\% of cells; we do not
claim universal isomorph discrimination. \textbf{Component~K: PASS.}

\subsection{Warm-Start Mechanism}
\label{sec:c2}

The oracle ablation replaces cold-prediction memory writes with
ground-truth reachability writes. This is non-deployable --- we do not
have ground truth at memory-write time in practice --- but it measures
what the inference mechanism achieves in this finite-memory setting
given perfect stored values. The result is not a global upper bound;
finite memory coverage, retrieval mismatch on the held-out validation
set, top-$K$ averaging, and sampler stochasticity all remain potential
limits even in the oracle condition. We report both warm-start variants from Methods: \textbf{Option~B}
at $t_{\mathrm{inject}}{=}2$ and \textbf{Option~A} at
$K_{\mathrm{refine}}{=}10$. The two paths are independent ---
$K_{\mathrm{refine}}$ does not appear in Option~B,
$t_{\mathrm{inject}}$ does not appear in Option~A.

Two observations follow from Table~\ref{tab:warm}'s oracle arm.
First, the \textbf{load-bearing lift is in $\mathrm{rec}_{-}$}: it
moves from 0.505
(chance) cold to 0.960 under Option-B warm inference with ground-truth
memory; $\mathrm{rec}_{+}$ stays at $\approx 0.99$. The mechanism is
correcting the negative class, not inflating accuracy via prior
collapse. Second, both variants beat the $2\times$ speedup gate
($3.3\times$ Option~B, $5.5\times$ Option~A) at near-saturated balanced
accuracy.

A $\beta$ sweep ($\beta \in \{20, 40, 80, 160\}$, both variants ---
eight oracle cells total) passes the gate by $\ge +18.91$\,pp in every
case. Notable: with oracle memory, \emph{lower} $\beta$ slightly
outperforms higher $\beta$ (Option~B $\Delta$bal at $\beta{=}20$:
$+22.39$\,pp; at $\beta{=}160$: $+20.75$\,pp), inverting the
cold-pred-memory trend (Section~\ref{sec:c3}). Interpretation:
high-quality stored values benefit from top-$K$ averaging that smooths
cell-level disagreement between near-correct neighbours; low-quality
stored values prefer peaky $\beta$ to concentrate on the least-bad
neighbour. \textbf{Optimal $\beta$ depends on memory quality, not an
architectural constant.} \textbf{Component~M: PASS.}

\subsection{Stored-Value Quality}
\label{sec:c3}

\begin{figure}[t]
\centering
\includegraphics[width=\columnwidth]{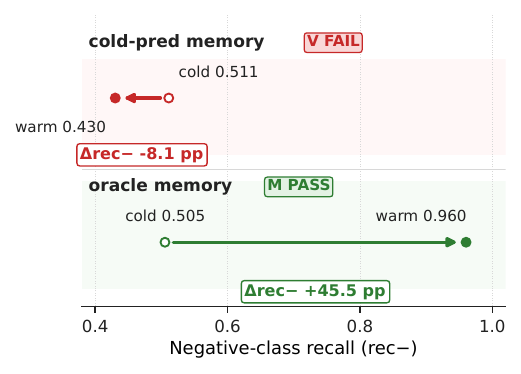}
\caption{\textbf{Negative-class recall carries the connectivity-2
story} (Option~B, $t_{\mathrm{inject}}{=}2$, $\beta{=}20$).
Cold-prediction memory \emph{degrades} $\mathrm{rec}_{-}$ under
warm-start ($0.511 \rightarrow 0.430$); oracle memory recovers it
sharply ($0.505 \rightarrow 0.960$). $\mathrm{rec}_{+}$ stays near
$1.0$ throughout.}
\label{fig:f5}
\end{figure}

Replacing oracle memory writes with cold-prediction memory writes is
the only change from Section~\ref{sec:c2} --- same key encoder,
warm-start dispatch, retrieval temperature, and validation set. Both
warm arms come from the same run so the comparison is direct; the two
cold rows reflect within-run sampling jitter ($\approx 0.3$\,pp here;
the cold noise floor measured separately across 5 seeds is balanced
accuracy std $0.13$\,pp). All $\Delta$s reported here and in
Section~\ref{sec:decompose-lift} dominate the noise floor by $\ge
11\times$ (smallest, per-query uniform) to $\ge 170\times$ (oracle).

The cold-pred warm row (Table~\ref{tab:warm}, sanity arm) misses the $-2$\,pp gate by
$2.09$\,pp. More informatively, \textbf{$\mathrm{rec}_{-}$ gets worse,
not better}: $0.511$ cold $\rightarrow 0.430$ warm in the same arm
(Figure~\ref{fig:f5}). The warm path inherits and amplifies the cold
model's class bias rather than failing to fix it; retrieval finds
high-similarity neighbours (key quality is fine), but every neighbour's
stored prediction is itself biased toward the positive class.

The oracle/cold-pred gap ($\Delta$bal $= +22.39$\,pp vs $-4.09$\,pp
within the same run) is the cleanest component-attribution evidence we
have: $26.5$\,pp of balanced accuracy separates \emph{identical}
mechanism and retrieval, differing only in stored-value quality. An
independent earlier run reproduces the cold-pred result at $\Delta$bal
$= -3.75$\,pp on a different cold draw, qualitatively similar (both
miss the $-2$\,pp gate by comparable margins). By the
decomposition of Section~\ref{sec:decomp}, the mechanism and retrieval
components are not the bottleneck on this task; the stored-value
component is. \textbf{Component~V: FAIL.}

\subsection{Bias Shift Versus Alignment: The Five-Arm Suite}
\label{sec:decompose-lift}

\begin{figure}[t]
\centering
\includegraphics[width=\columnwidth]{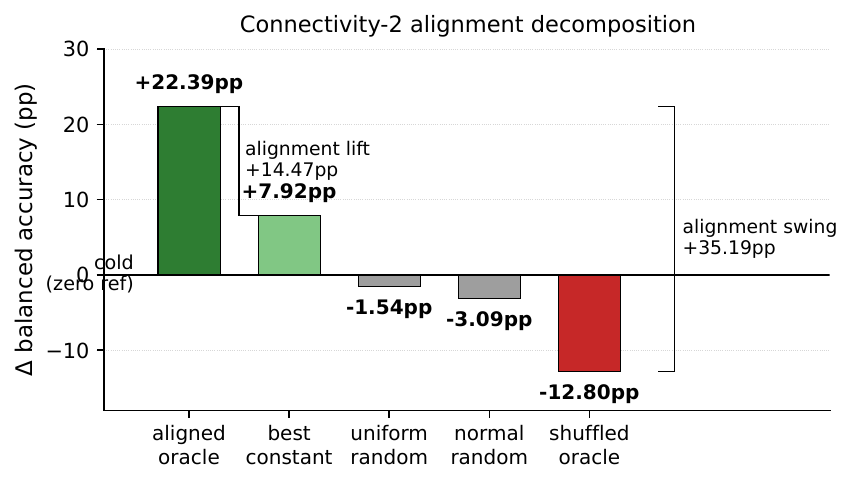}
\caption{\textbf{Alignment decomposition on connectivity-2} (Option~B,
$t_{\mathrm{inject}}{=}2$, $\beta{=}20$). The aligned oracle exceeds
the best bias-only warm-init by $+14.47$\,pp and a shuffled oracle
(same gold values, (key, value) pairings permuted across queries) by
$+35.19$\,pp, isolating per-graph alignment as the dominant lever.
Per-arm $\Delta$bal in Table~\ref{tab:decomp}.}
\label{fig:f1}
\end{figure}

\begin{table}[t]
\centering
\small
\begin{tabular}{lrr}
\toprule
comparison & Option~B & Option~A \\
\midrule
oracle aligned vs.\ cold              & $+22.39$ & $+20.45$ \\
best constant vs.\ cold               & $+7.92$  & $+5.65$  \\
per-query uniform $[-1,1]$ vs.\ cold  & $-1.54$  & $-10.25$ \\
per-query $\mathcal{N}(0,1)$ vs.\ cold & $-3.09$  & $-10.46$ \\
shuffled oracle vs.\ cold             & $-12.80$ & $-14.30$ \\
\midrule
\textbf{aligned $-$ best constant}    & \textbf{$+14.47$} & \textbf{$+14.80$} \\
\textbf{aligned $-$ shuffled}         & \textbf{$+35.19$} & \textbf{$+34.75$} \\
\bottomrule
\end{tabular}
\caption{Decomposition of the oracle lift, $\Delta$bal in pp at
$\beta{=}20$. Best constant: $c{=}-0.25$ (Option~B), $c{=}-0.75$
(Option~A).}
\label{tab:decomp}
\end{table}

Sections~\ref{sec:c2} and~\ref{sec:c3} contrasted the oracle
mechanism's $+22$\,pp gain with the deployable pipeline's failure. We
now scrutinise the oracle lift itself: is it genuine per-graph value
reuse, or class-prior shift alone?
Equivalently: would any sufficiently-negative warm init achieve a
similar lift by breaking the cold model's positive-bias equilibrium,
with retrieval contributing nothing? Three ablations decompose it: a
constant warm-init sweep $c \in \{-1, \ldots, 1\}$; a
\textbf{shuffled-oracle} arm (gold values, but with (key, value)
pairings permuted across queries --- preserving the value distribution
while destroying per-graph alignment); and a \textbf{per-query
stochastic init} arm (self-retrieve with per-entry random storage at
$\beta{=}20$, $t_{\mathrm{inject}}{=}2$; uniform $[-1, 1]$ and
$\mathcal{N}(0, 1)$).

The constant sweep stores a constant tensor of value $c$ at every
memory slot; retrieval over identical values is a no-op, so the
warm-start receives exactly $c$. The per-query stochastic arms use
self-retrieve with per-entry random storage --- top-1 self-similarity
dominates the softmax (mean retrieval-top weight $0.887$), so each
query retrieves predominantly its own pre-stored random vector.
Table~\ref{tab:decomp} and Figure~\ref{fig:f1} report the results.

The decomposition tells a consistent story. Bias shift is bounded: no
constant warm init exceeds $+8$\,pp $\Delta$bal. On top of that best
bias-only baseline, aligned ground-truth retrieval adds $+14.5$\,pp,
and the two warm-start variants --- which consume the init
differently --- agree on this gap to within $0.4$\,pp. Misalignment
does not merely fail to help but actively poisons: the shuffled oracle
falls $\approx 20$\,pp \emph{below} the best constant, so alignment is
a dominant, non-additive lever rather than a small perturbation. And
per-query stochasticity is not the missing axis --- both random-init
arms fall at or below cold under Option~B.

We state the contribution precisely. That lift is the
\emph{aligned-vs-best-constant} gap, not a one-sided ``retrieval
contributes $+14.5$\,pp'' claim; the full alignment effect is the
$+35$\,pp swing from misaligned to aligned ground truth. A tuned
constant prior delivers up to $+8$\,pp; the remaining lift requires
per-graph alignment between key and stored value, which contributes
nearly twice what the best bias shift does. The objection that
the oracle merely supplies a better class prior or a generic warm-init
regulariser does not survive these controls.

\subsection{Why Stored-Value Quality Fails}
\label{sec:why}

\begin{figure}[t]
\centering
\includegraphics[width=\columnwidth]{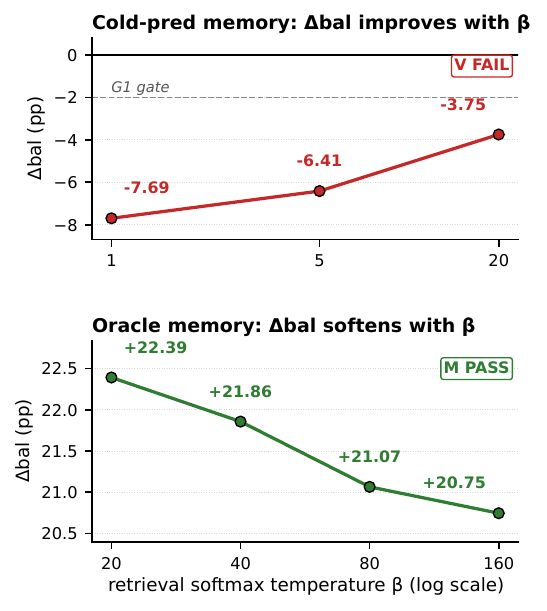}
\caption{\textbf{$\beta$-trend inversion: $\Delta$bal's response to
$\beta$ depends on stored-value quality} (Option~B,
$t_{\mathrm{inject}}{=}2$). Cold-prediction memory: $\Delta$bal
improves monotonically with $\beta$ but never clears the $-2$\,pp
gate. Oracle memory: trend inverts --- highest at the lowest $\beta$
tested, softening as $\beta$ grows. Non-overlapping $\beta$-grids;
diagnostic trends, not a head-to-head sweep.}
\label{fig:f4}
\end{figure}

Two interventions further characterise the failure as representational
rather than procedural.

\paragraph{Threshold tuning.} Cold raw outputs lie in $[-0.99, +0.99]$
and are evaluated at threshold $\tau{=}0$. Sweeping $\tau \in [-1, 1]$
in steps of $0.02$ over the validation set and selecting $\tau$ for
balanced accuracy yields a maximum lift of $+0.47$\,pp (at
$\tau{=}+0.96$: acc $0.821$, bal.\ acc $0.760$, $\mathrm{rec}_{+}$
$1.000$, $\mathrm{rec}_{-}$ $0.520$, against $\tau{=}0$: $0.817$,
$0.755$, $1.000$, $0.511$), small relative to the $-2$\,pp gate and
not qualitatively corrective. Crucially, $\mathrm{rec}_{+}$ stays at $1.000$ up to
$\tau{=}+0.96$ --- the model does not even consider assigning negative
class until the threshold reaches 96\% of its prediction range. Threshold shifting reveals no recoverable balanced-accuracy fix
under this probe; features are sharply committed and largely
class-biased. The ``right features, wrong decision boundary''
diagnosis is not supported.

\paragraph{Class-weighted retraining.} A more invasive intervention:
modify IRED's denoising MSE to apply a $4\times$ per-cell weight on
negative-class cells, by subclassing the IRED denoising objective with
a per-cell loss weight applied inside the diffusion loss, mirroring
the existing per-cell weighting precedent in IRED's shortest-path
task. After 30{,}000 steps under the same
schedule, balanced accuracy \emph{decreased} by $3.2$\,pp: the original
G0 reaches bal.\ acc $0.755$ ($\mathrm{rec}_{+}$ $1.000$,
$\mathrm{rec}_{-}$ $0.511$ --- strongly biased toward the positive
class), while the rebalanced G0 reaches $0.724$ ($\mathrm{rec}_{+}$
$0.447$, $\mathrm{rec}_{-}$ $1.000$ --- flips toward the negative
class). The
retrained model did not find a balanced equilibrium; it flipped
polarity. Two degenerate equilibria with no balanced middle suggest the
learned representation does not support balanced classification
under the tested training recipe --- class weighting decides only
which side training collapses to.

The two probes converge on the same conclusion: the stored-value limit
is consistent with a learned-representation limit under this training
recipe rather than with a tunable boundary or loss weight. Two interventions that \emph{should} have helped if the failure were
only a decision-threshold or class-weighting artifact both fail to. The $\beta$-trend inversion of
Figure~\ref{fig:f4} is the same story read through retrieval
temperature: the direction of the trend depends on stored-value
quality.

Several questions remain open. We have not shown that a larger or
differently-architected G0 --- a deeper GNN, an attention-based
variant, an alternative loss --- could not solve connectivity-2, nor
that the oracle ceiling generalises to tasks where the cold model is
already strong. By extension we expect iterative refill of warm
predictions into memory to amplify cold bias, but we have not tested
this formally. Our reverse path is the stochastic DDPM
\texttt{p\_sample}; whether a deterministic-reverse DDIM path behaves
identically is untested --- Option~B covers the forward-init half, not
the reverse. Finally, the warm-start ablations run on a single fixed
validation set, and multi-seed warm-start replication is left for a
larger study.

\section{Sudoku: A Key-Quality Failure}
\label{sec:sudoku}

\begin{figure*}[t]
\centering
\includegraphics[width=\textwidth]{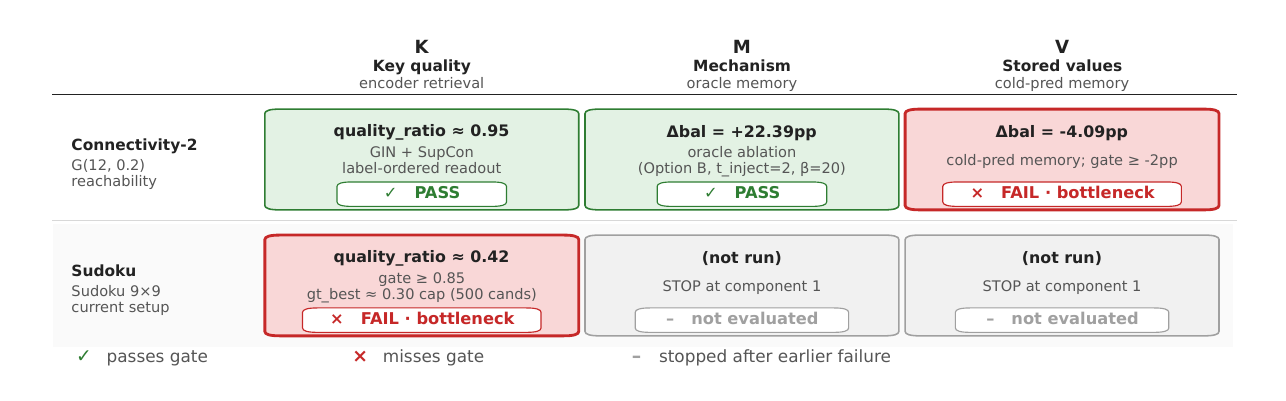}
\caption{\textbf{Heterogeneous failure modes across two reasoning
tasks} (connectivity rows: Option~B, $t_{\mathrm{inject}}{=}2$,
$\beta{=}20$; Sudoku: key-quality screen only). On connectivity-2, K
and oracle M pass; the deployable cold-pred pipeline then fails at V
($-4.09$\,pp vs the $-2$\,pp gate). On Sudoku, the encoder itself
blocks at K ($\texttt{quality\_ratio} \approx 0.42$ vs the $\ge 0.85$
gate, $\texttt{gt\_best} \approx 0.30$ candidate-pool ceiling).}
\label{fig:f3}
\end{figure*}

\begin{table}[t]
\centering
\small
\begin{tabular}{lrrr}
\toprule
metric & gate & best (step) & step 3000 \\
\midrule
\texttt{quality\_ratio}              & $\ge 0.85$ & 0.434 (2000) & 0.420 \\
$\texttt{ret\_top\_w}(\beta{=}20)$   & $\ge 0.30$ & 0.242 (3000) & 0.242 \\
\texttt{recall@1}                    & diagnostic & 0.010 (1500) & 0.006 \\
loss EMA                             & ($\approx 4.16$) & --- & 5.290 \\
\bottomrule
\end{tabular}
\caption{Sudoku key-encoder training (3000 SupCon steps). Both
key-quality kill criteria fail by margin, and validation metrics
plateau despite continued loss descent; we stop before evaluating
M/V.}
\label{tab:sudoku}
\end{table}

Applying the same diagnostic logic to Sudoku --- stopped at the
key-quality screen, before the warm-start mechanism and stored-value
components are exercised --- produces a clean negative at a
\emph{different} component than connectivity. The ResNet key encoder
(Methods) trained 3{,}000 steps reaches \texttt{quality\_ratio}
$0.42$ (gate $\ge 0.85$, missed by half) and
$\texttt{ret\_top\_w}(\beta{=}20)$ $0.242$ (gate $\ge 0.30$, missed by
$0.058$ absolute); both metrics plateau (Table~\ref{tab:sudoku}). Loss
descends but validation metrics decouple after step 1500; we do not
proceed to warm-start diagnostics under this setup.

\begin{figure}[!ht]
\centering
\includegraphics[width=\columnwidth]{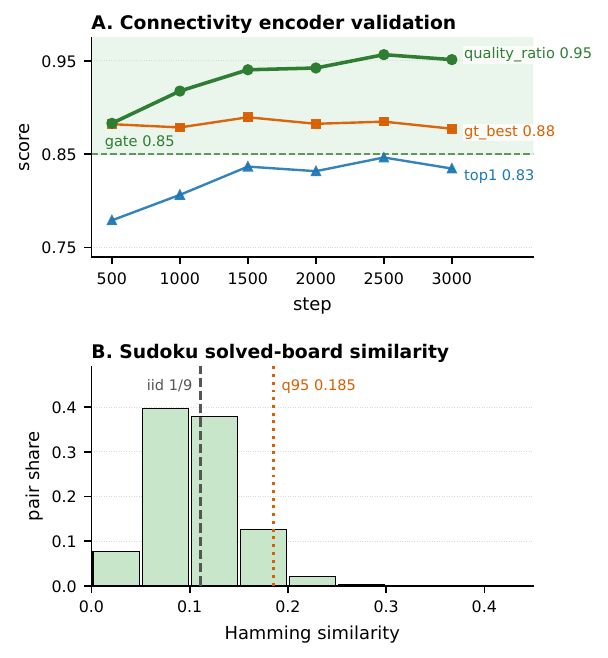}
\caption{(A) Connectivity-2 encoder validation:
\texttt{quality\_ratio} clears the $\ge 0.85$ gate by step 500,
saturating near $0.95$ against the per-batch
\texttt{gt\_best} $\approx 0.88$ ceiling. (B) Sudoku solved-board
Hamming similarity (999{,}000 pairs): unimodal at iid $1/9$,
q95 $= 0.185$, \texttt{gt\_best} $\approx 0.30$ --- the candidate-pool
side of Sudoku's K-screen failure (see Section~\ref{sec:sudoku} for
the encoder-side limit).}
\label{fig:f6}
\end{figure}

The Hamming-similarity distribution on solved boards
(Figure~\ref{fig:f6}B) is unimodal at the iid baseline (mean $= 1/9
\approx 0.111$; q95 $= 0.185$, max $= 0.395$). \texttt{gt\_best} ---
the in-batch maximum target similarity per anchor under the mask-aware
target across 500 candidates --- plateaus at \textbf{$\approx 0.30$}
across all training checkpoints. Even with a perfect encoder, only
$\approx 30\%$ of the query's unknown cells would match the true
digit --- a weak warm-start target relative to the cold endpoint's
$0.97$ cell accuracy. The downstream warm-start effect is not
measured here: Sudoku stops at the K-screen, before M/V are
exercised.

Two distinct limits hold under the current setup. The encoder
under-retrieves: \texttt{quality\_ratio} $0.42$ means top-1 retrievals
reach less than half the in-batch best available similarity, so the
encoder retrieves substantially worse than the candidate pool allows.
Separately, the in-batch best itself plateaus at \texttt{gt\_best}
$\approx 0.30$ --- a candidate-pool / target ceiling, not an encoder
verdict, so even a perfect retriever delivers warm-starts at only
$\approx 0.30$ absolute target similarity to the query. A better encoder alone could in principle clear the normalized
\texttt{quality\_ratio} gate; producing a useful warm-start signal on
Sudoku, however, requires moving both limits. Sudoku fails at
component~K (key quality, under the current setup); connectivity
fails at component~V (stored-value quality, with the mechanism
working) --- Figure~\ref{fig:f3}. A symmetry-aware encoder, a
canonicalised or larger candidate pool, or a different value target
could move these limits; we did not run those configurations.

\section{Related Work}
\label{sec:related}

RW-EBR wraps IRED \cite{du2024ired}, whose energy-based diffusion
solves the connectivity-2 and Sudoku tasks studied here; IRED's
ablations cover gradient-on-energy, multi-step refinement, and
contrastive shaping --- not memory or warm-start initialisation. IREM
\cite{du2022irem} is the earlier energy-minimisation ancestor, also
without memory or warm-start. The warm-start step itself ---
forward-noise a retrieved value to $t_{\mathrm{inject}}$, then denoise
from $t_{\mathrm{inject}}$ down to 0 --- is the standard partial-noise
warm-start introduced by SDEdit \cite{meng2022sdedit} for image
editing; the forward step is the standard diffusion marginal $q(x_t
\mid x_0)$ shared by DDPM \cite{ho2020ddpm} and DDIM
\cite{song2021ddim}, with DDIM distinguished by its deterministic
reverse sampler (we use the stochastic DDPM \texttt{p\_sample}). The
closest learned-warm-start competitor is WSD \cite{scholz2025wsd},
which learns a per-query informed Gaussian prior for conditional
generation. Sampler-distillation approaches such as Consistency Models
\cite{song2023consistency} are a complementary acceleration axis
(few-step approximation rather than warm-start init), not exercised
here.

RDM \cite{blattmann2022rdm} and kNN-Diffusion
\cite{sheynin2022knndiffusion} condition image diffusion models on
retrieved nearest neighbours; MEMENTO \cite{chalumeau2025memento}
conditions a combinatorial-optimisation solver on a memory bank. In
all of these the retrieved item enters as a conditioning signal
alongside the input; RW-EBR differs in the role of retrieval, where
the retrieved vector is the warm-start initialisation of the sampler's
dynamical update rather than a conditioning vector. The trajectory
memory itself is a Modern Hopfield network
\cite{ramsauer2021hopfield}; a recent theoretical line
\cite{ambrogioni2024,pham2025} bridges diffusion models and
associative memory, which we use operationally --- retrieve a
candidate trajectory and feed it as a warm-start --- rather than
developing it theoretically.

RAGChecker \cite{ru2024ragchecker} and RAG-X \cite{ragx2026} decompose
LLM retrieval-augmented-generation pipelines into retriever- and
generator-side diagnostic metrics for failure attribution. Our
three-component decomposition (key quality / warm-start mechanism /
stored-value quality) adapts this logic to retrieval-warmed
\emph{iterative inference}; we do not claim the decomposition, the
partial-noise warm-start mechanism, or the oracle-memory upper bound
as novel primitives. The slot that differs from the LLM-RAG setting
is the warm-start mechanism --- the retrieved vector enters a
dynamical update, not a context window. What is distinctive across
these prior threads is the application: we are not aware of any that
tests its warm-start against a constant baseline, a per-query random
control, and a shuffled-oracle arm, or that reports a component-level
decomposition across two reasoning tasks under iterative inference.

\section{Conclusion}

We introduced a five-arm ablation methodology --- oracle,
best-constant, per-query-random, shuffled, and aligned --- for
attributing the behaviour of retrieval-warmed iterative inference.
The full suite is exercised on connectivity-2 through an IRED
energy-based diffusion backbone augmented with a Modern Hopfield
trajectory memory; the same diagnostic logic is applied to Sudoku as
a key-quality screen.
On connectivity-2 the suite isolates per-graph key-value alignment as
the dominant lever: a $+35$\,pp aligned-vs-shuffled-oracle swing that a
tuned constant prior and per-query stochastic initialisation together
cannot explain. The same decomposition identifies the first blocking component of
the deployable pipeline on each task: stored-value quality on
connectivity-2, key quality on Sudoku under the current setup. The oracle
arms are diagnostic upper bounds, not deployable results; their value
is in attribution. Coverage beyond these two tasks remains future work. Where a single end-to-end
accuracy number reports only that retrieval warm-starting helped or
failed, the methodology names the component --- key quality,
warm-start mechanism, or stored-value quality --- that carried it: a
localised verdict no aggregate score can give.

\bibliographystyle{named}
\bibliography{references}

\end{document}